\documentclass{article}
\pdfoutput=1
\PassOptionsToPackage{numbers, compress, sort}{natbib}

\usepackage[preprint]{neurips_2024}
\usepackage{hyperref}  

\usepackage[utf8]{inputenc} 
\usepackage[T1]{fontenc}    
\usepackage{url}            
\usepackage{booktabs}       
\usepackage{amsfonts}       
\usepackage{nicefrac}       
\usepackage{microtype}      
\usepackage{xcolor}         
\usepackage{graphicx}
\usepackage{subcaption}
\usepackage{wrapfig}
\usepackage{xcolor}
\usepackage{tcolorbox}
\usepackage{float}

\title{Unpacking the Individual Components of Diffusion Policy}

\author{%
Xiu Yuan\\
University of California, San Diego \\
\texttt{x1yuan@ucsd.edu}\\
}

\begin{document}

\maketitle

\begin{abstract}
Imitation Learning presents a promising approach for learning generalizable and complex robotic skills. The recently proposed Diffusion Policy generates robot action sequences through a conditional denoising diffusion process, achieving state-of-the-art performance compared to other imitation learning methods. This paper summarizes five key components of Diffusion Policy: 1) observation sequence input; 2) action sequence execution; 3) receding horizon; 4) U-Net or Transformer network architecture; and 5) FiLM conditioning. By conducting experiments across ManiSkill and Adroit benchmarks, this study aims to elucidate the contribution of each component to the success of Diffusion Policy in various scenarios. We hope our findings will provide valuable insights for the application of Diffusion Policy in future research and industry.
\end{abstract}

\section{Introduction}

\begin{wrapfigure}{r}{0.50\linewidth}
    \centering
    \vspace{-0.4cm}
    \includegraphics[width=\linewidth]{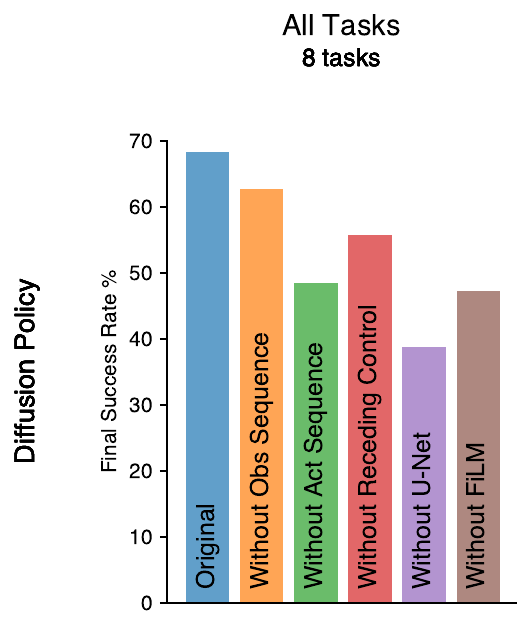}
    \vspace{-0.6cm}
    \caption{Relative importance of individual components across 8 tasks on ManiSkill and Adroit.}
    \label{fig:bar_all}
    \vspace{-0.5cm}
\end{wrapfigure}

Imitation learning provides an efficient approach for teaching robots to perform various complex tasks, such as grasping \citep{johns2021coarse, xie2020deep, stepputtis2020language}, legged locomotion \citep{ratliff2007imitation, al2023locomujoco}, dexterous manipulation \citep{qin2022dexmv, radosavovic2021state}, and mobile manipulation \citep{wong2022error, du2022bayesian}. With advancements in computer vision and natural language processing, increasingly sophisticated imitation learning architectures have been developed, demonstrating impressive performance across diverse tasks \citep{chen2021decision, abramson1970aloha, florence2022implicit, shafiullah2022behavior}. Recently, Diffusion Policy, proposed by \cite{chi2023diffusion}, introduced an innovative approach by representing robot action sequences through a conditional denoising diffusion process, achieving state-of-the-art performance compared to other imitation learning methods. As a result, Diffusion Policy has gained popularity and is now widely used as a policy backbone in research and downstream applications.

Despite Diffusion Policy's popularity, a key question remains unanswered: how does each of its components contribute to overall performance? As Diffusion Policy gains traction within the robotics community, we observe that many researchers, either intentionally or unintentionally, modify its structure for their own purposes—often without thorough examination of each component. For example, \cite{ren2024diffusion} uses a single observation frame rather than a stack of past frames and executes all predicted actions without receding horizon control. They also employ a Multi-Layer Perceptron (MLP) as the denoising network architecture in their primary experiments. Similarly, \cite{ze20243d} predicts and executes only four actions per inference, compared to the 16 used in Diffusion Policy's original design. These variations highlight an urgent need for a systematic summary of Diffusion Policy’s essential components and a comprehensive study illustrating how these elements affect performance in different scenarios. This information would provide researchers with a clearer foundation for considering potential component modifications to optimize their applications.

\begin{wrapfigure}{r}{0.50\linewidth}
    \centering
    \includegraphics[width=\linewidth]{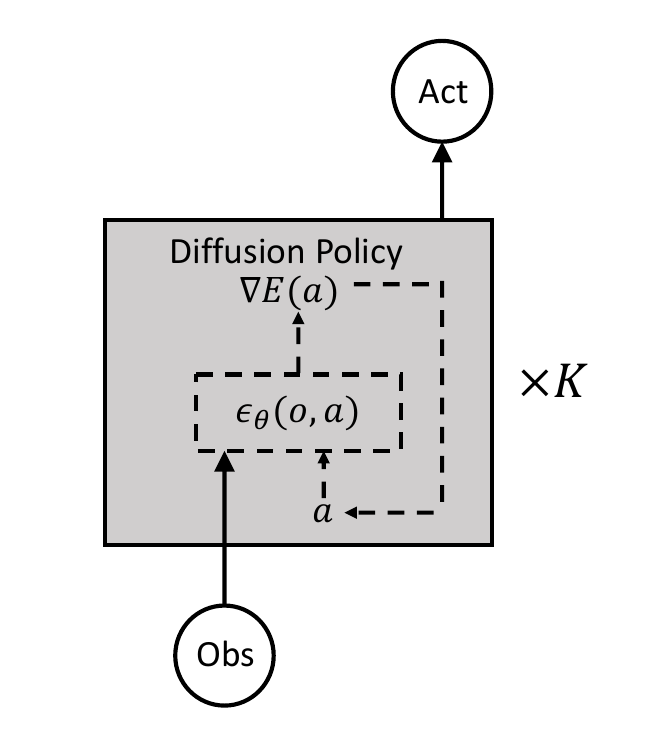}
    \vspace{-0.6cm}
    \caption{Overview of Diffusion Policy}
    \label{fig:dp}
\end{wrapfigure}

In this study, we summarize 5 key components of Diffusion Policy:

\begin{itemize}
    \item Diffusion Policy takes a stack of past observation sequences instead of one frame of current observations. We call it \textbf{observation sequence input}
    \item Diffusion Policy executes a sequence of actions in the environment at one inference instead of only executing one action. We call it \textbf{action sequence execution}.
    \item Diffusion Policy predicts many actions at one inference but only execute the first few actions in the environment instead of executing all actions predicted. We call it \textbf{receding horizon control}.
    \item Diffusion Policy employs an U-Net \citep{ronneberger2015u} or Transformer architecture \citep{vaswani2017attention} as denoising network backbone instead of MLP. We call it \textbf{denoising network architecture}.
    \item  Diffusion Policy takes observations as FiLM conditioning \citep{perez2018film} instead of taking them as network input. We call it \textbf{FiLM conditioning}.
\end{itemize}

After identifying and summarizing the five key components of Diffusion Policy, we conduct ablation experiments on each component using ManiSkill and Adroit benchmarks. These experiments aim to reveal the individual contribution of each component to the overall performance of Diffusion Policy. We conclude our findings as follows:

\begin{itemize}
    \item \textbf{Observation sequence input}: Observation sequence input is crucial for tasks requiring Absolute Control, but has little impact on tasks in Delta Control Mode, where a single observation is sufficient.
    \item \textbf{Action sequence execution}: Action Sequence Execution improves performance by 10-20\% for most tasks, but for tasks requiring real-time control, such as the Adroit Hammer task, shorter action horizons or single action roll-outs are preferred due to their responsiveness.
    \item \textbf{Receding horizon control}: Receding Horizon Control enhances performance for long horizon tasks but has little effect on short horizon tasks, as it is designed to optimize long-term planning.
    \item \textbf{Denoising network architecture}: U-Net denoising architecture is crucial for hard tasks, while MLP denoising architecture is sufficient for easy tasks, emphasizing the need to choose the appropriate network based on task complexity.
    \item \textbf{FiLM conditioning}: FiLM Conditioning greatly improves the performance of the Diffusion Policy on hard tasks, but is not necessary for easy tasks.
\end{itemize}

In light of these observations, we offer the following recommendations for future research and applications of Diffusion Policy:

\begin{itemize}
    \item \textbf{Observation Sequence Input}: se observation sequence input for tasks requiring Absolute Control, but a single observation is sufficient for Delta Control tasks.
    \item \textbf{Action Sequence Execution}: Apply action sequence execution for most tasks, but for tasks requiring real-time control (e.g., Adroit Hammer), prefer shorter action horizons or single action roll-outs for better responsiveness. 
    \item \textbf{Receding Horizon Control}: Implement receding horizon control for long horizon tasks to optimize long-term planning, but it is unnecessary for short horizon tasks.
    \item \textbf{Denoising Network Architecture}: Use U-Net denoising architecture for hard tasks and MLP denoising architecture for easy tasks, depending on task complexity
    \item \textbf{FiLM Conditioning}: Apply FiLM conditioning to enhance performance on hard tasks, but it is not required for easy tasks.
\end{itemize}

\section{Related Works}

\textbf{Diffusion model as policy} With the success of diffusion models in image synthesis and video generation \citep{ho2020denoising}, they have become a popular choice as policy backbones in the robotics community. These models are utilized in two main ways: 1) As policies in reinforcement learning (RL) methods, including offline RL \citep{wang2022diffusion} \citep{hansen2023idql} \citep{mao2024diffusion}, offline-to-online RL \citep{ding2023consistency}, and online RL \citep{yang2023policy}; 2) As policies in imitation learning \citep{chi2023diffusion} \citep{reuss2023goal}. Diffusion Policy belongs to the second category and has demonstrated state-of-the-art performance compared to other imitation learning methods \citep{shafiullah2022behavior} \citep{florence2022implicit} \citep{abramson1970aloha}. Furthermore, it exhibits significant potential for future research and practical applications. For these reasons, we have chosen Diffusion Policy as the primary focus of our study.

\section{Diffusion Policy}

Diffusion Policy \citep{chi2023diffusion} is a sophisticated system that leverages a conditional denoising diffusion process at its core to generate actions, along with several additional techniques and design choices, as shown in Fig. \ref{fig:dp}. We systematically categorize these into five key components, which are thoroughly detailed in the following sections.

\subsection{Observation Sequence Input}

\begin{wrapfigure}{r}{0.50\linewidth}
    \vspace{-0.3cm}
    \centering
    \includegraphics[width=\linewidth]{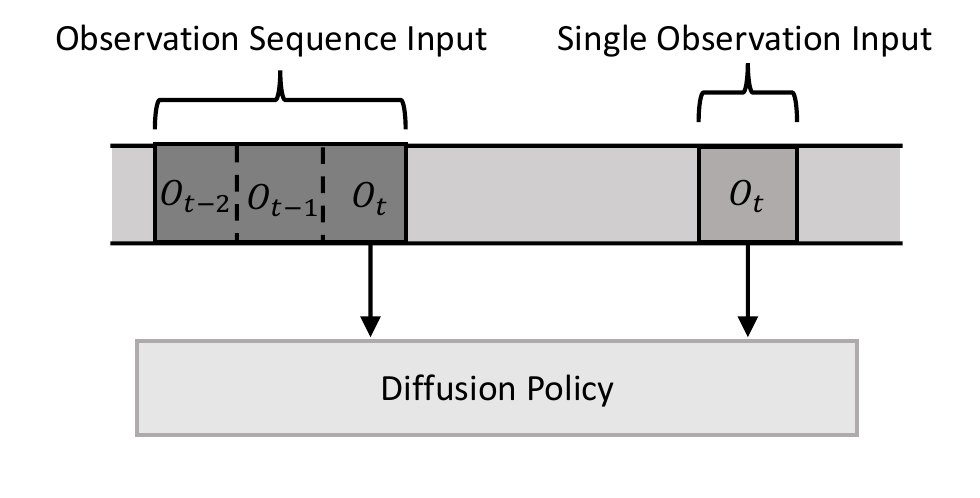}
    \vspace{-0.6cm}
    \caption{Visualization of observation sequence input and single observation input}
    \label{fig:obs_input}
    \vspace{-0.6cm}
\end{wrapfigure}

Many recent imitation learning models, such as BeT \cite{shafiullah2022behavior} and ACT \cite{abramson1970aloha}, are history-dependent. In other words, these models utilize a sequence of past observations as input, rather than relying solely on the current observation. As highlighted in \cite{mandlekar2021matters}, data—particularly human demonstrations—often depend on contextual information spanning multiple observations. Similarly, Diffusion Policy adheres to this history-dependent paradigm. As illustrated in Fig. \ref{fig:obs_input}, Diffusion Policy formally takes as input a window of past observations, $[O_t, O_{t-1}, ..., O_{t-T_o}]$, where $T_o$ denotes the \textbf{observation horizon}.

\subsection{Action Sequence Execution}

\begin{wrapfigure}{r}{0.50\linewidth}
    \vspace{-1.6cm}
    \centering
    \includegraphics[width=\linewidth]{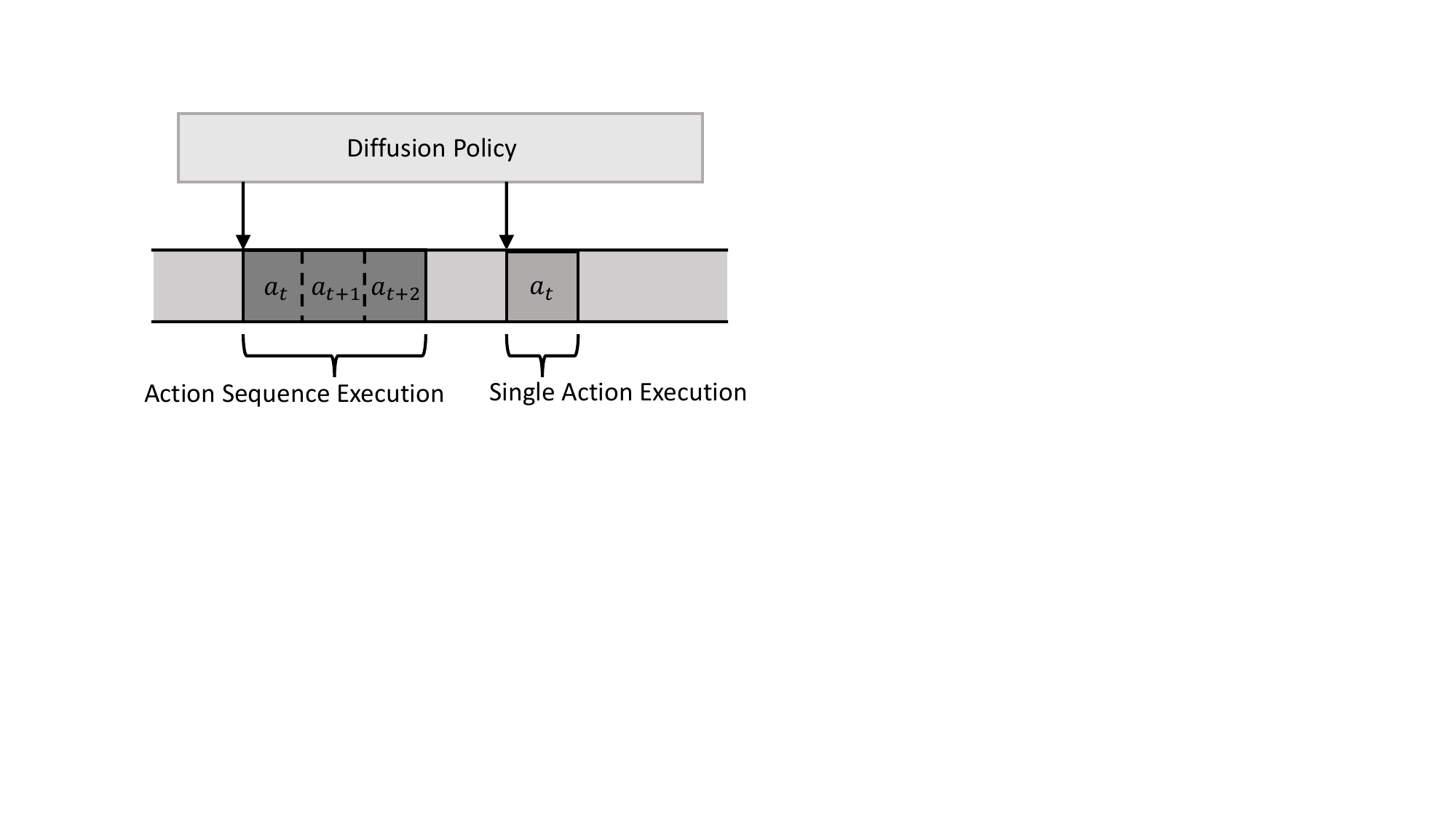}
    \vspace{-0.6cm}
    \caption{Visualization of action sequence execution and single action execution}
    \label{fig:act_seq}
\end{wrapfigure}

Unlike most imitation learning models, which execute a single action at each inference step (e.g., \cite{shafiullah2022behavior}, \cite{florence2022implicit}), Diffusion Policy adopts a different strategy by executing multiple subsequent actions in the environment. As demonstrated in Fig. \ref{fig:act_seq}, Diffusion Policy executes $[a_t, a_{t+1, ..., a_{t+T_a}}]$, where $T_a$ denotes the \textbf{action horizon}.

\subsection{Receding Horizon Control}

\begin{wrapfigure}{r}{0.90\linewidth}
    \centering
    \vspace{-0.8cm}
    \includegraphics[width=\linewidth]{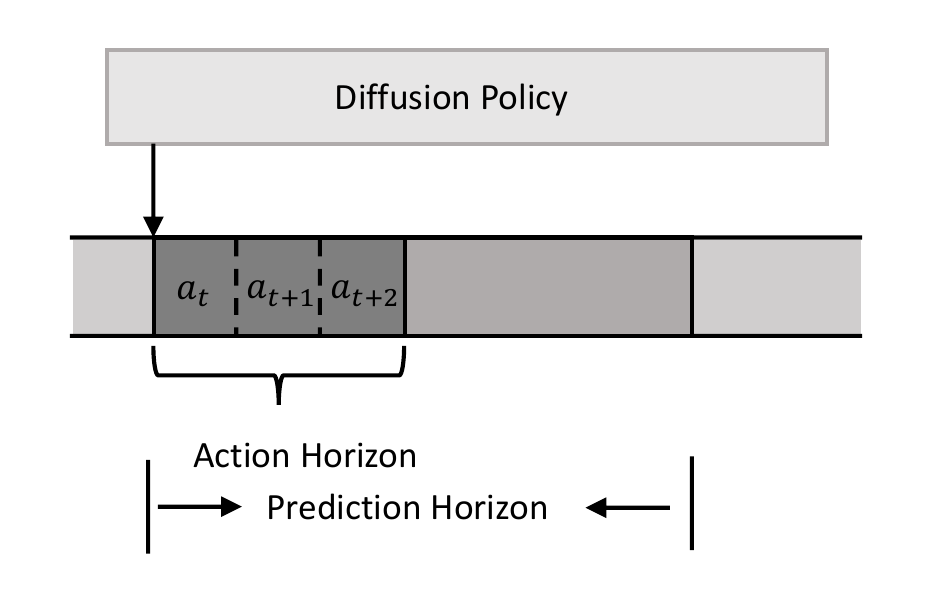}
    \vspace{-0.6cm}
    \caption{Visualization of receding horizon control}
    \label{fig:receding}
\end{wrapfigure}

Diffusion Policy innovatively employs \textbf{receding horizon control} aiming to strike a balance between re-generating actions in time and maintaining temporal action consisteny. Formally, at one inference time, Diffusion Policy predicts $[a_t, a_{t+1}, ..., a_{t+T_p}]$ subsequent actions and only select the first $[a_t, a_{t+1}, ..., a_{t+T_a}]$ to roll out in the environment, with $T_a < T_p$, as illustrated by Fig. \ref{fig:receding}.

\subsection{Denoising Network Architecture}

The architecture of the denoising network is crucial for diffusion models to achieve high performance. In \cite{ho2020denoising}, a U-Net \cite{ronneberger2015u}-based denoiser was introduced and subsequently became a standard in diffusion models for image synthesis and video generation. Later, \cite{peebles2023scalable} proposed a transformer-based denoiser that demonstrated stronger performance, albeit with less stable training, and it has been adopted in many state-of-the-art image generation works, such as \cite{leedalle} and \cite{saharia2022photorealistic}. However, for a considerable time, MLPs were predominantly used as the denoising network architecture in diffusion-based online and offline RL works. It was not until the introduction of Diffusion Policy that U-Net and transformer-based denoisers gained popularity within the robotics community.

\subsection{FiLM Conditioning}

Most imitation learning methods \cite{shafiullah2022behavior} \cite{abramson1970aloha} \cite{mandlekar2021matters} \cite{florence2022implicit} directly use the observation or observation sequence as the input to their networks. In contrast, Diffusion Policy draws inspiration from vision domains, treating the observation sequence as FiLM conditioning on the denoising network.

\begin{wrapfigure}{r}{0.40\linewidth}
    \centering
    \includegraphics[width=\linewidth]{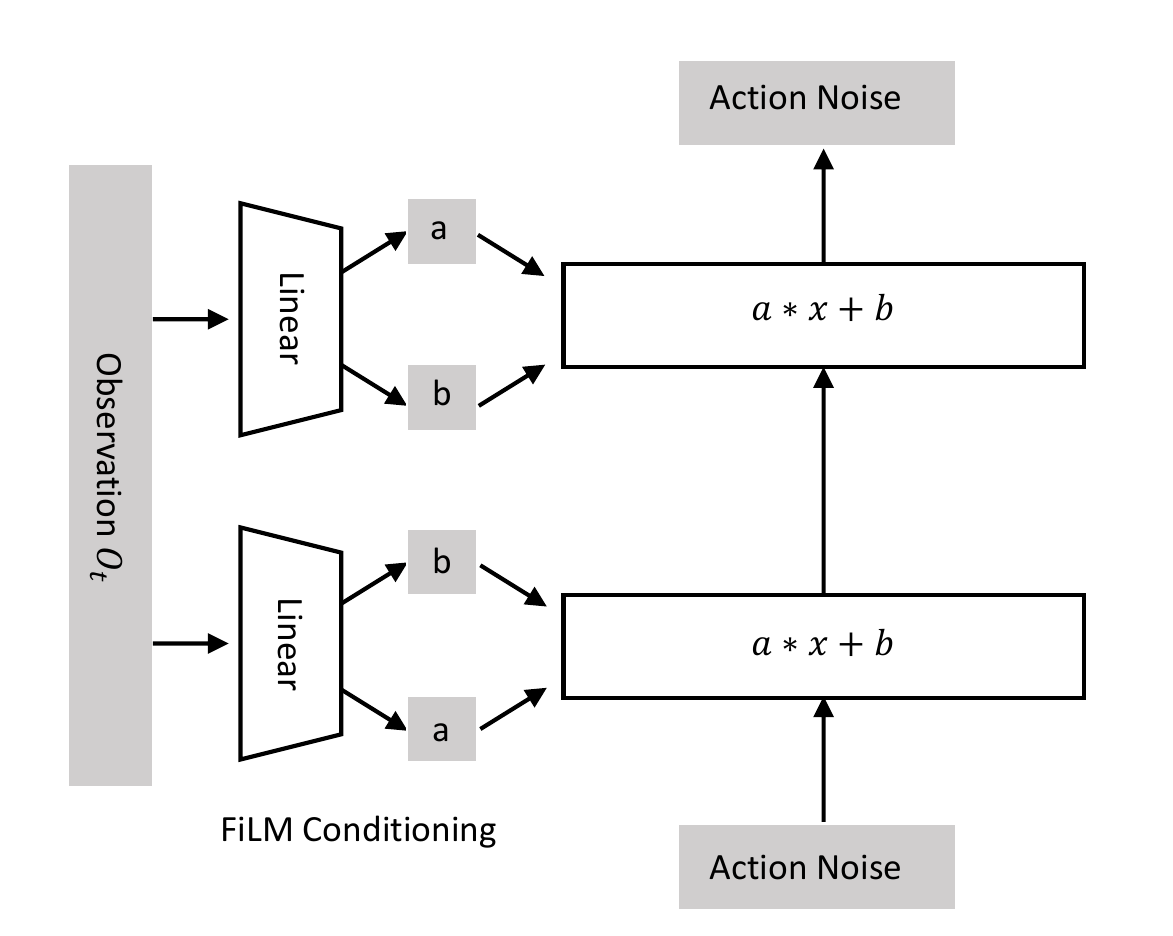}
    \vspace{-0.6cm}
    \caption{Visualization of FiLM conditioning}
    \label{fig:film}
\end{wrapfigure}

\section{Experiments}

The goal of our experimental evaluation is to study the following questions:

\begin{enumerate}
    \item How does \textbf{Observation Sequence Input} contribute to the performance of Diffusion Policy (Sec. \ref{sec:obs_seq})?
    \item How does \textbf{Action Sequence Execution} contribute to the performance of Diffusion Policy (Sec. \ref{sec:act_seq})?
    \item How does \textbf{Receding Horizon Control} contribute to the performance of Diffusion Policy (Sec. \ref{sec:receding})?
    \item How does \textbf{Denoising Network Architecture} contribute to the performance of Diffusion Policy (Sec. \ref{sec:denoising})?
    \item How does \textbf{FiLM Conditioning} contribute to the performance of Diffusion Policy (Sec. \ref{sec:film})?
\end{enumerate}

\subsection{Experimental Setup}

\begin{figure}[t]
    \centering
    \vspace{-0.5 cm}
    \includegraphics[width=0.99\textwidth]{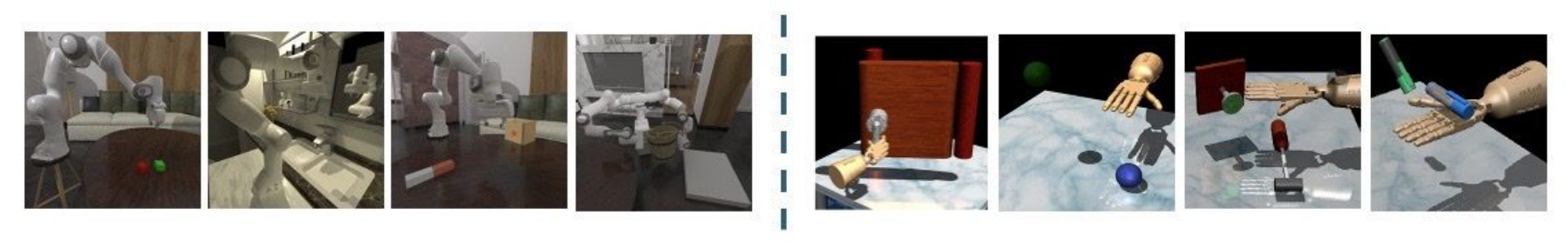}
    \vspace{-0.2 cm}
    \caption{\textbf{Tasks Visualizations.} ManiSkill (left four figures) and Adroit (right four figures).} 
    \vspace{-0.5 cm}
    \label{fig:task_vis}
\end{figure}

Our experimental setup incorporates \textbf{variations across the following dimensions}:

\begin{itemize}
    \item \textbf{Benchmarks}: ManiSkill and Adroit.
    \item \textbf{Task Types}: Stationary robot arm manipulation, mobile manipulation, dual-arm coordination, dexterous hand manipulation, articulated object manipulation, and high-precision tasks. Fig. \ref{fig:task_vis} illustrates sample tasks from each benchmark.
    \item \textbf{Demonstration Sources}: Task and Motion Planning (TAMP), Model Predictive Control (MPC), Reinforcement Learning, and Human Demonstrations.
    \item \textbf{Observation Modalities}: low-dimensional state observation.
\end{itemize}

\subsubsection{Task Description}

Our experiments are conducted on 8 tasks across 2 benchmarks: ManiSkill (robotic manipulation; 4 tasks), and Adroit (dexterous manipulation; 4 tasks).
See Fig. \ref{fig:task_vis} for illustrations.



\textbf{ManiSkill} 
We consider four challenging tasks from ManiSkill. StackCube and PegInsertionSide demand \textit{high-precision control}, with PegInsertion featuring a mere 3mm clearance. TurnFaucet and PushChair introduce \textit{object variations}, where the base policy is trained on source environment objects, but target environments for online interactions contain different objects.
For all ManiSkill tasks, we use 1000 demonstrations provided by the benchmark \citep{mu2021maniskill} \citep{gu2023maniskill2} across all methods. These demonstrations are generated through task and motion planning, model predictive control, and reinforcement learning. 


\textbf{Adroit} We consider all four dexterous manipulation tasks from Adroit: Door, Hammer, Pen, and Relocate. The tasks should be solved using a complex, 24-DoF manipulator, simulating a real hand. 
For all Adroit tasks, we use 25 demonstrations provided by the original paper \citep{rajeswaran2017learning} for all methods. These demonstrations are collected by human teleoperation.

\subsection{Observation Sequence Input}
\label{sec:obs_seq}

\begin{table}[h]
\centering
\caption{The performance of Diffusion Policy with and without \textbf{Observation Sequence Input}.}
\begin{tabular}{lccccl}  
\toprule
Task & Control Mode & With & Without \\
\midrule
ManiSkill: StackCube & Delta Control & \textbf{99\%} & \textbf{98\%}\\
ManiSkill: PegInsertionSide & Delta Control & \textbf{80\%} & \textbf{81\%} \\
ManiSkill: TurnFaucet & Delta Control & \textbf{59\%} & \textbf{55\%} \\
ManiSkill: PushChair & Absolute Control & \textbf{61\%} & 55\%\\
Adroit: Door & Absolute Control & \textbf{95\%} & 83\% \\
Adroit: Pen & Absolute Control & \textbf{71\%} & \textbf{72\%}\\
Adroit: Hammer & Absolute Control & \textbf{17\%}& 11\% \\
Adroit: Relocate & Absolute Control & \textbf{64\%}& 47\%\\
\bottomrule
\label{tab:obs_seq}
\end{tabular}
\end{table}

\begin{figure}[h]
    \centering
    \includegraphics[width=0.45\textwidth]{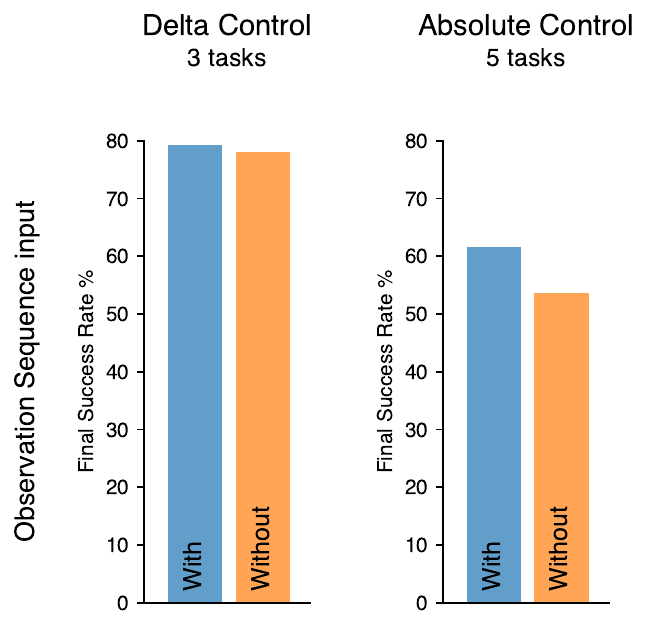}
    \vspace{-0.3 cm}
    \caption{
        Performance comparison of Diffusion Policy with observation sequence input and without observation sequence input under delta control and absolute control.
    }
    \label{fig:bar_obs_seq}
\end{figure}

In this section, we aim to understand how 
\textbf{Observation Sequence Input} affects the performance of the Diffusion Policy. We evaluated the Diffusion Policy with and without observation sequence input across 8 tasks from the ManiSkill and Adroit benchmarks, as shown in Tab. \ref{tab:obs_seq}. Among these tasks, three are in \textbf{Delta Control Mode}, which requires the policy to output the delta change in the robot's position, pose, or velocity, either in the world frame or the end-effector frame. The remaining five tasks are in \textbf{Absolute Control Mode}, where the policy must output the absolute values of the robot's position, pose, or velocity.

As illustrated in Fig. \ref{fig:bar_obs_seq}, removing the observation sequence input leads to an overall 10\% performance drop for tasks requiring Absolute Control, while having little impact on tasks that rely on Delta Control.

The empirical results in Tab. \ref{tab:obs_seq} and Fig. \ref{fig:bar_obs_seq} suggest that past observations are essential for inferring the robot's movements, orientation, and changes in position relative to the environment. In contrast, for tasks with Delta Control, a single observation frame suffices, as only relative changes need to be considered.

\begin{tcolorbox}[colframe=blue!40!black, colback=blue!10!white, coltitle=cyan!50!white, fonttitle=\bfseries, title=Takeaway 4.2: Observation Sequence Input]
Observation sequence input significantly impacts the performance of Diffusion Policy in tasks requiring Absolute Control. However, for tasks in Delta Control Mode, the performance remains largely unaffected. This suggests that past observations are crucial for tasks requiring absolute control inference, but a single observation is sufficient for delta control tasks.
\end{tcolorbox}

\subsection{Action Sequence Execution}
\label{sec:act_seq}

\begin{table}[h]
\centering
\caption{The performance of Diffusion Policy with and without \textbf{Action Sequence Execution}.}
\begin{tabular}{lcccl}  
\toprule
Task & With & Without \\
\midrule
ManiSkill: StackCube & \textbf{99\%}& 86\% \\
ManiSkill: PegInsertionSide & \textbf{80\%}& 55\% \\
ManiSkill: TurnFaucet & \textbf{59\%}& 35\%\\
ManiSkill: PushChair & \textbf{61\%}& 46\%\\
Adroit: Door & \textbf{95\%}& 46\%\\
Adroit: Pen & \textbf{71\%}& 58\%\\
Adroit: Hammer & 17\% & \textbf{27\%} \\
Adroit: Relocate & \textbf{64\%}& 35\%\\
\bottomrule
\label{tab:act_seq}
\end{tabular}
\end{table}

\begin{figure}[h]
    \centering
    \includegraphics[width=0.45\textwidth]{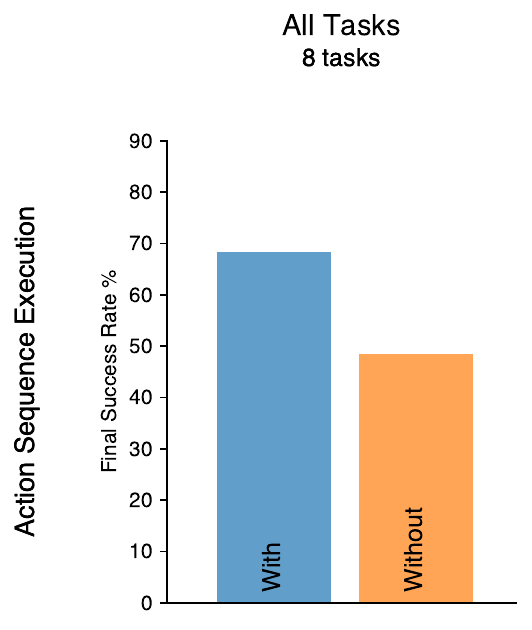}
    \vspace{-0.3 cm}
    \caption{
        Performance comparison of Diffusion Policy with action sequence execution and without action sequence execution under all tasks.
    }
    \label{fig:bar_act_seq}
\end{figure}

In this section, we aim to understand how Action Sequence Execution affects the performance of the Diffusion Policy. We evaluated the Diffusion Policy with and without action sequence execution across 8 tasks from the ManiSkill and Adroit benchmarks, as shown in Tab. \ref{tab:act_seq}.

The empirical results in Tab. \ref{tab:act_seq} and Fig. \ref{fig:act_seq} indicate that, for most tasks, action sequence execution provides a consistent performance boost of 10-20\%. One exception is the Adroit Hammer task, where single action execution outperforms action sequence execution due to its more responsive actions. A direct roll-out of 8 subsequent actions loses the ability to receive real-time feedback from the environment, thus diminishing responsiveness. The key takeaway is that for tasks requiring responsive control and real-time environment feedback, shorter action horizons or even single action roll-outs are preferred

\begin{tcolorbox}[colframe=blue!40!black, colback=blue!10!white, coltitle=cyan!50!white, fonttitle=\bfseries, title=Takeaway 4.3: Action Sequence Execution]
Action Sequence Execution generally improves performance by 10-20\% for most tasks. However, for tasks like the Adroit Hammer task, single action execution is more effective due to its responsiveness, as longer action sequences lose the ability to respond to real-time feedback. Therefore, for tasks requiring real-time control, shorter action horizons or single action roll-outs are preferred.
\end{tcolorbox}

\subsection{Receding Horizon Control}
\label{sec:receding}

\begin{table}[h]
\centering
\caption{The performance of Diffusion Policy with and without \textbf{Receding Horizon Control}.}
\begin{tabular}{lccccl}  
\toprule
Task & Task Horizon/Truncation Steps & With & Without\\
\midrule
ManiSkill: StackCube & 140/200 &\textbf{99\%} & 88\% \\
ManiSkill: PegInsertionSide & 160/200 &\textbf{80\%} & 72\% \\
ManiSkill: TurnFaucet & 140/200 &\textbf{59\%} & 50\% \\
ManiSkill: PushChair & 130/200 &\textbf{60\%} & 51\% \\
Adroit: Door & 180/300 &\textbf{95\%}& 85\% \\
Adroit: Pen & 30/200 &\textbf{71\%} & \textbf{73\%} \\
Adroit: Hammer & 270/400 &\textbf{17\%}& \textbf{17\%} \\
Adroit: Relocate & 270/400 &\textbf{64\%}& 11\% \\
\bottomrule
\label{tab:receding}
\end{tabular}
\end{table}

\begin{figure}[h]
    \centering
    \includegraphics[width=0.45\textwidth]{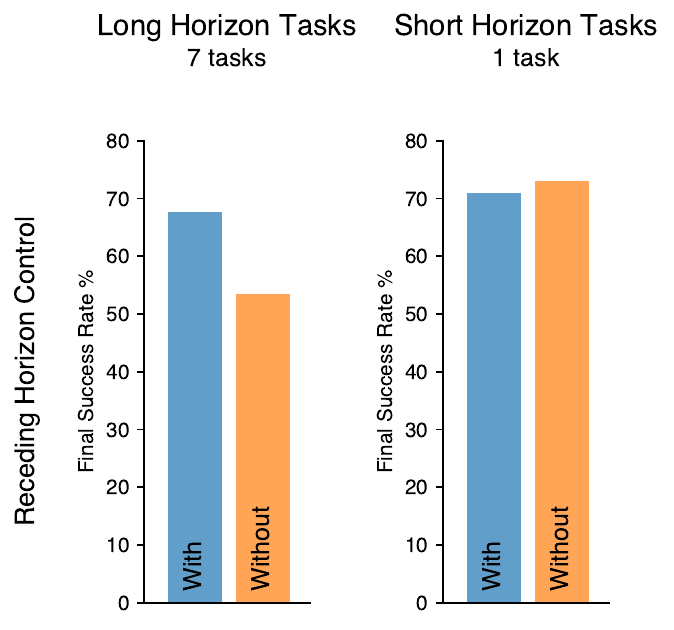}
    \vspace{-0.3 cm}
    \caption{
        Performance comparison of Diffusion Policy with receding horizon control and without receding horizon control under long horizon tasks and short horizon tasks.
    }
    \label{fig:bar_receding}
\end{figure}

In this section, we aim to understand how Receding Horizon Control affects the performance of the Diffusion Policy. We evaluated the Diffusion Policy with and without receding horizon control across 8 tasks from the ManiSkill and Adroit benchmarks, as shown in Tab. \ref{tab:receding}. The task horizon, calculated as the average number of steps required to successfully complete each task, is presented in the table. Seven tasks have a task horizon greater than 100 and are classified as long horizon tasks, while one task has a task horizon of 30, classified as a short horizon task.

As shown in Tab. \ref{tab:receding} and Fig. \ref{fig:bar_receding}, removing receding horizon control results in approximately a 15\% performance drop for long horizon tasks, while causing a slight performance increase for short horizon tasks.

The empirical results indicate that receding horizon control is essential for long horizon tasks but unnecessary for short horizon tasks. Since receding horizon control is specifically designed to enhance the capability for long-term planning, it provides little benefit for tasks that require only a few steps to complete.

\begin{tcolorbox}[colframe=blue!40!black, colback=blue!10!white, coltitle=cyan!50!white, fonttitle=\bfseries, title=Takeaway 4.4: Receding Horizon Control]
Receding Horizon Control significantly improves performance for long horizon tasks but has minimal impact on short horizon tasks. This is because receding horizon control is designed to optimize long-term planning, making it essential for tasks requiring many steps, while offering little advantage for tasks that are completed in just a few steps.
\end{tcolorbox}

\subsection{Denoising Network Architecture}
\label{sec:denoising}

\begin{table}[h]
\centering
\caption{The performance of Diffusion Policy with different \textbf{Denoising Network Architecture}.}
\begin{tabular}{lccccl}  
\toprule
Task & Task Difficulty & \textbf{U-Net} & \textbf{MLP}\\
\midrule
ManiSkill: StackCube & Easy &\textbf{99\%} & \textbf{99\%} \\
ManiSkill: PegInsertionSide & Hard &\textbf{80\%} & 21\% \\
ManiSkill: TurnFaucet & Hard &\textbf{59\%} & 22\%\\
ManiSkill: PushChair & Hard &\textbf{60\%} & 42\%\\
Adroit: Door & Hard &\textbf{95\%} & 35\% \\
Adroit: Pen & Easy &\textbf{71\%}& \textbf{68\%}\\
Adroit: Hammer & Hard &\textbf{17\%} & \textbf{17\%}\\
Adroit: Relocate & Hard &\textbf{64\%}& 7\% \\
\bottomrule
\label{tab:denoising}
\end{tabular}
\end{table}

\begin{figure}[h]
    \centering
    \includegraphics[width=0.45\textwidth]{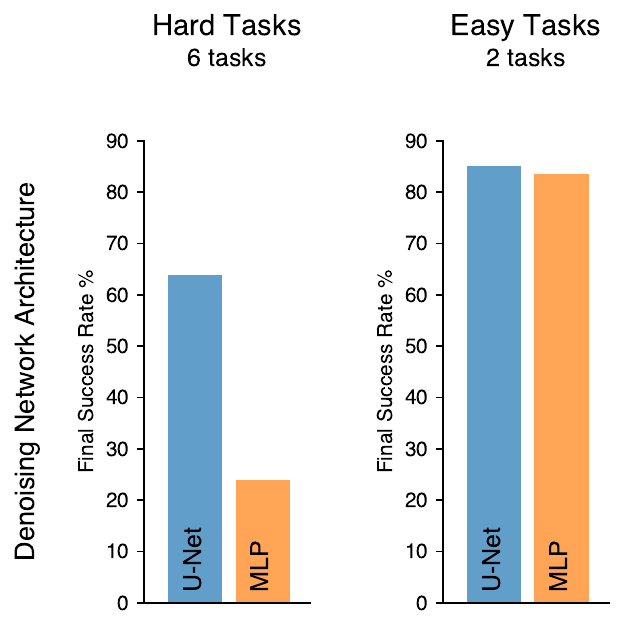}
    \vspace{-0.3 cm}
    \caption{
        Performance comparison of Diffusion Policy with U-Net and MLP under hard tasks and easy tasks.
    }
    \label{fig:bar_denoising}
\end{figure}

In this section, we aim to understand how the \textbf{Denoising Network Architecture} affects the performance of the Diffusion Policy. We evaluated the Diffusion Policy with both U-Net and MLP denoising networks across 8 tasks from the ManiSkill and Adroit benchmarks, as shown in Tab. \ref{tab:denoising}. The tasks are categorized into easy and hard tasks based on their inherent difficulty. StackCube (ManiSkill) is considered an easy task due to its simple pick-and-place nature with object variations. Pen (Adroit) is classified as an easy task because it requires a very short task horizon. The remaining tasks—such as TurnFaucet, PushChair, PegInsertionSide, Door, Hammer, and Relocate—involve challenges such as object variations, the need for precise control, and lower quality demonstrations, and are thus classified as hard tasks.

The empirical results in Tab. \ref{tab:denoising} and Fig. \ref{fig:bar_denoising} indicate that the U-Net denoising architecture is crucial for achieving strong performance on hard tasks, while the MLP denoising architecture is sufficient for easy tasks.

\begin{tcolorbox}[colframe=blue!40!black, colback=blue!10!white, coltitle=cyan!50!white, fonttitle=\bfseries, title=Takeaway 4.5: Denoising Network Architecture]
U-Net denoising architecture is essential for achieving strong performance on hard tasks, while the MLP denoising architecture is sufficient for easy tasks. This highlights the importance of choosing the appropriate denoising network based on task complexity.
\end{tcolorbox}

\subsection{FiLM Conditioning}
\label{sec:film}

\begin{table}[h]
\centering
\caption{The performance of Diffusion Policy with \textbf{FiLM Conditioning} and \textbf{Observation as Direct Inputs}.}
\begin{tabular}{lccccl}  
\toprule
Task & Task Difficulty & \textbf{FiLM Conditioning} & \textbf{Direct Inputs}\\
\midrule
ManiSkill: StackCube & Easy &\textbf{99\%} & \textbf{97\%}\\
ManiSkill: PegInsertionSide & Difficult &\textbf{80\%} & 44\%\\
ManiSkill: TurnFaucet & Difficult &\textbf{59\%} & 27\%\\
ManiSkill: PushChair & Difficult &\textbf{60\%} & 36\%\\
Adroit: Door & Difficult &\textbf{95\%} & 79\% \\
Adroit: Pen & Easy &\textbf{71\%} & \textbf{75\%}\\
Adroit: Hammer & Difficult &\textbf{17\%}& 18\% \\
Adroit: Relocate & Difficult &\textbf{64\%}& 2\% \\
\bottomrule
\label{tab:film}
\end{tabular}
\end{table}

\begin{figure}[h]
    \centering
    \includegraphics[width=0.45\textwidth]{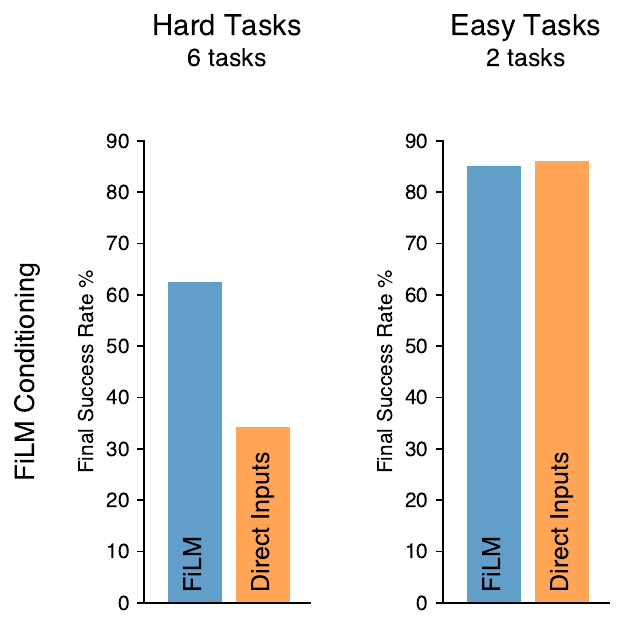}
    \vspace{-0.3 cm}
    \caption{
        Performance comparison of Diffusion Policy with FiLM conditioning and direct inputs under hard tasks and easy tasks.
    }
    \label{fig:bar_film}
\end{figure}

In this section, we aim to understand how FiLM Conditioning affects the performance of the Diffusion Policy. We evaluated the Diffusion Policy with FiLM conditioning and direct inputs across 8 tasks from the ManiSkill and Adroit benchmarks, as shown in Tab. \ref{tab:film}.

The empirical results in Tab. \ref{tab:film} and Fig. \ref{fig:bar_film} indicate that FiLM conditioning significantly improves performance on hard tasks, while it is not necessary for easy tasks.

\begin{tcolorbox}[colframe=blue!40!black, colback=blue!10!white, coltitle=cyan!50!white, fonttitle=\bfseries, title=Takeaway 4.6: FiLM Conditioning]
FiLM Conditioning significantly enhances the performance of the Diffusion Policy on hard tasks, but is not needed for easy tasks.
\end{tcolorbox}

\section{Conclusions}
In this study, we systematically decompose the Diffusion Policy into five distinct components: 1) Observation Sequence Input, 2) Action Sequence Execution, 3) Receding Horizon Control, 4) Denoising Network Architecture, and 5) FiLM Conditioning. We evaluate the relative importance of each component using the ManiSkill and Adroit benchmarks and provide recommendations for researchers and practitioners.

\newpage
\bibliographystyle{plain}
\bibliography{ref}

\newpage
\appendix



\end{document}